\documentclass[11pt]{article} 
\usepackage{url}
\usepackage{smile}
\usepackage{CJKutf8}
\usepackage{graphicx} 
\usepackage{todonotes}
\usepackage{epstopdf}
\usepackage{wrapfig}
\usepackage[colorlinks, linkcolor=blue, anchorcolor=blue, citecolor=blue]{hyperref}
\usepackage[margin=1in]{geometry}
\usepackage[normalem]{ulem}
\usepackage[export]{adjustbox}
\usepackage{mathtools, cuted}
\usepackage{natbib}
\usepackage{enumerate}
\usepackage{microtype}
\usepackage{graphicx}
\usepackage{booktabs} 
\usepackage{wrapfig}
\usepackage{xspace}

\usepackage{kpfonts}
\DeclareMathAlphabet{\mathsf}{OT1}{cmss}{m}{n}
\SetMathAlphabet{\mathsf}{bold}{OT1}{cmss}{bx}{n}

\begin{document}

\title{Multi-Domain Neural Machine Translation with Word-Level Adaptive Layer-wise Domain Mixing}

\author{Haoming Jiang, Chen Liang, Chong Wang, and Tuo Zhao  \thanks{Haoming Jiang, Chen Liang, and Tuo Zhao are affiliated with Georgia Institute of Technology.Chong Wang is affiliated with ByteDance.  Emails: \texttt{\{hmjiang,cliang73\}@gatech.edu} \texttt{chong.wang@bytedance.com} \texttt{tourzhao@gatech.edu}. }}

\date{}

\maketitle


\begin{abstract}
	
	Many multi-domain neural machine translation (NMT) models achieve knowledge transfer by enforcing one encoder to learn shared embedding across domains. However, this design lacks adaptation to individual domains. To overcome this limitation, we propose a novel multi-domain NMT model using individual modules for each domain, on which we apply word-level, adaptive and layer-wise domain mixing.
	We first observe that words in a sentence are often related to multiple domains. Hence, we assume each word has a domain proportion, which indicates its domain preference. Then word representations are obtained by mixing their embedding in individual domains based on their domain proportions.
	We show this can be achieved by carefully designing multi-head dot-product attention modules for different domains, and eventually taking weighted averages of their parameters by word-level layer-wise domain proportions.
	Through this, we can achieve effective domain knowledge sharing, and capture fine-grained domain-specific knowledge as well. Our experiments show that our proposed model outperforms existing ones in several NMT tasks.
	
\end{abstract}

\vspace{-0.1in}
\section{Introduction} \label{sec:intro}
\vspace{-0.1in}

Neural Machine Translation (NMT) has made significant progress in various machine translation tasks \citep{kalchbrenner2013recurrent,sutskever2014sequence,bahdanau2014neural,luong2015effective,wu2016google}. 
The success of NMT heavily relies on a huge amount of annotated parallel sentences as training data, which is often limited in certain domains, e.g., medical domain.  One approach to address this is to explore unparalleled corpora, such as unsupervised machine translation~\citep{lample2017unsupervised,lample2018phrase}. Another approach is to train a multi-domain NMT model and this is the focus of this paper. The simplest way is to build a unified model by directly pooling all training data from multiple domains together, as the languages from different domains often share some similar semantic traits, e.g., sentence structure, textual style and word usages. For domains with less training data, the unified model usually shows significant improvement. 

Researchers have proposed many methods for improving multi-domain NMT. Though certain semantic traits are shared across domains, there still exists significant heterogeneity among languages from different domains. For example, \citet{haddow2012analysing} show that for a domain with sufficient training data, a unified model may lead to weaker performance than the one trained solely over the domain; \citet{farajian2017neural,luong2015effective,sennrich2015improving,servan2016domain} also show that to improve the translation performance over certain domains, fine-tuning the unified model is often needed, but at the expense of sacrificing the performance over other domains. This indicates that a unified model might not well exploit the domain-specific knowledge for each individual domain. 

To overcome this drawback, two lines of recent research focus on developing new methods by exploiting domain-shared and domain-specific knowledge to improve multi-domain NMT \citep{britz2017effective,zeng2018multi,tars2018multi,hashimoto2016domain, wang2017instance,chen2017cost,wang2018sentence}.  

One line of research focuses on instance weighting, which assigns domain related weights to different samples during training. For example, \citet{wang2017instance} consider sentence weighting and domain weighting for NMT. The sentence weight is determined by the bilingual cross-entropy of each sentence pair based on the language model of each domain. The domain weight can be modified by changing the number of sentences from that domain in a mini-batch. \citet{chen2017cost} propose a cost weighting method, where the weight of each pair of sentences is evaluated by the output probability of a domain classifier on the encoder embedding. \citet{wang2018sentence} propose a dynamic training method to adjust the sentence selection and weighting during training. We remark that many of these methods are complementary to our proposed model, and can be applied to improve the training of our model.

Another line of research attempts to design specific encoder-decoder architectures for NMT models. For example, \citet{britz2017effective} consider domain-aware embedding given by the encoder, and then jointly train a domain classifier, taking the embedding as input to incorporate the domain information. \citet{zeng2018multi} further extend their approach by separating the domain-shared and domain-specific knowledge within the embedding. In addition, \citet{zeng2018multi} and \citet{shen2018word} propose a maximum weighted likelihood estimation method, where the weight is obtained by word-level domain aware masking to encourage the model to pay more attention to the domain-specific words. The aforementioned methods, however, have a notable limitation: They enforce one single encoder to learn shared embedding across all domains, which often lacks adaptivity to each individual domain. 

%
%
%
%
%
%



To better capture domain-shared knowledge beyond shared embedding from a single encoder, we propose a novel multi-domain NMT model using individual modules for each domain, on which we apply word-level, adaptive and layer-wise domain mixing. 
Our proposed model is motivated by the observation that although every sentence of the training data has a domain label, the words in the sentence are not necessarily only related to that domain. For instance, the word ``article'' appears in the domains of laws and business. Therefore, we expect the knowledge for translating the word ``article'' to be shared between these two domains. 
Our proposed model assigns a context-dependent domain proportion\footnote{A word actually has multiple domain proportions at different layers of our model. See more details in Section 3} to every word in the sentence. The domain proportions of the words can be naturally integrated into the Transformer model for capturing domain-shared/specific knowledge, as the multi-head dot-product attention mechanism is applied at the word-level. Specifically, we carefully design multi-head dot-product attention modules for different domains, and eventually mix these modules by taking weighted averages of their parameters by their layer-wise domain proportions.

Compared with existing models, ours has the following two advantages: 

\smallskip

\noindent $\bullet$ Our proposed model is more powerful in capturing the domain-specific knowledge, as we design multiple dot-product attention modules for different domains. In contrast, existing models rely on one single shared encoder, and then one single unified translation model is applied, which often cannot adapt to each individual domain very well.

\smallskip

\noindent $\bullet$ Our proposed model is more adaptive in the process of domain knowledge sharing. For common words across domains, their domain proportions tend to be uniform, and therefore can significantly encourage knowledge sharing. For some words specific to certain domains, their domain proportions tend to be skewed, and accordingly, the knowledge sharing is encouraged only within the relevant domains. For example, the word ``article'' appears less in the medical domain than the domains of laws and business. Therefore, the corresponding domain proportion tends to favor the domains of laws and business more than the medical domain.

\smallskip

We evaluate our proposed model in several multi-domain machine translation tasks, and the empirical results show that our proposed model outperforms existing ones and improves the translation performance for all domains.

\vspace{-0.1in}
\section{Background} \label{sec:background}
\vspace{-0.1in}

\textbf{Neural Machine Translation} (NMT) directly models the conditional distribution of the translated sentence $\yb=(y_1,...,y_\ell)$ given a source sentence
$\xb=(x_1,...,x_\ell)$.\footnote{Here we assume that we have applied padding to all sentences, and therefore, they are all of the same length.}. The conditional probability density function $p(\yb|\xb)$ is parameterized by an encoder-decoder neural network: The encoder encodes the source sentence into a sequence of hidden representations $\cH(\xb) = (h_1, ..., h_n)$, and the decoder generates target sentence one token at a time using these intermediate representations. More specifically, the decoder usually contains a recursive structure for computing $p(y_t|y_{<t},\xb)$ by
\begin{align*}
p(y_t|y_{<t},\xb)= \cF(\cG_t,\cH(\xb),y_{t-1}),
\end{align*}
where $\cG_{t}$ denotes the hidden representation of the decoder for the $t$-th position of the sequence, and $\cF$ denotes a multi-layered network that outputs the probability of $y_t$. Notice that $\cG_{t}$ is generated by the $\cG_{t-1}, \cH(\xb)$, and the previous word $y_{t-1}$. Given $N$ pairs of source/target sequences denoted by $\{\xb_i, \yb_i\}_{i=1}^n$, we train the NMT model by minimizing the cross-entropy loss as follows,
\begin{align*}
	\textstyle\min_{\cH,\cG,\cF}L_{\rm gen}= \frac{1}{n}\sum_{i=1}^n -\log p(\yb_i|\xb_i)
\end{align*}
where $p(\yb_i|\xb_i) = \prod_{t=1}^mp(y_{i,t}|y_{i,<t},\xb_i)$.

\medskip

\noindent\textbf{Transformer} is one of the most popular NMT models \citep{vaswani2017attention,tubay2018neural, devlin2018bert}. The encoder and decoder in Transformer contain stacked self-attention and point-wise, fully connected layers without any explicit recurrent structure, which is different from existing RNN-based NMT models. 


Specifically, \citet{vaswani2017attention} propose a new attention function using the scaled dot-product as the alignment score, which takes the form,
\begin{align}\label{dot-product}
\mathrm{Attention}(Q,K,V)=\mathrm{softmax}\Big(\frac{QK^\top}{\sqrt{d}}\Big)V,
\end{align}
where $Q,K,V\in \RR^{\ell \times d}$ are the vector representations of all the words in the sequences of queries, keys and values accordingly.  For the self-attention modules in the encoder and decoder, $Q=K=V$; For the attention module that takes into account the encoder and the decoder sequences, $Q$ is different from the sequence represented by $V$ and $K$.

\begin{figure}[!ht]
	\centering
	\vspace{-0.1in}
	\includegraphics[width=0.4\textwidth]{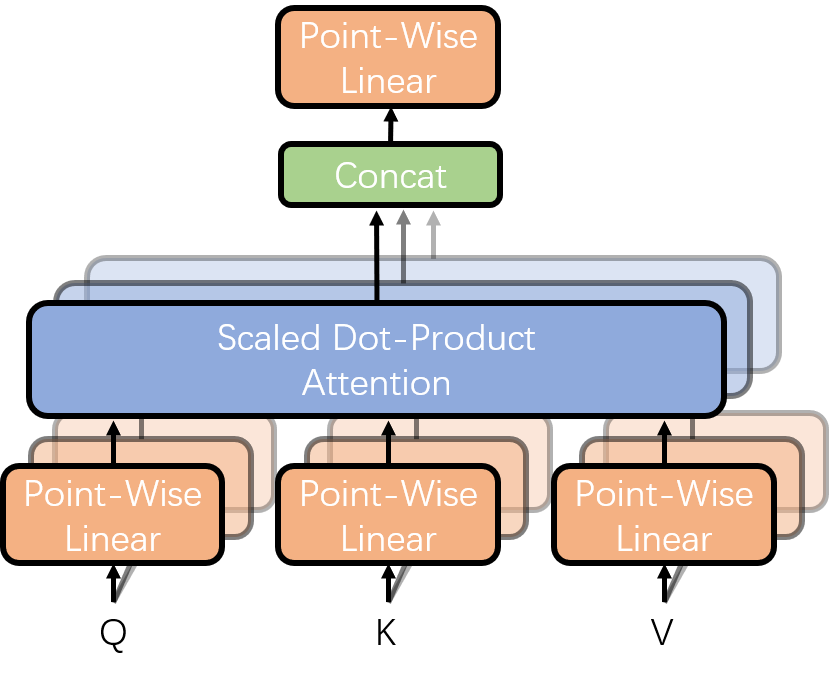}
	\vspace{-0.2in}
	\caption{Multi-head Scaled Dot-Product Attention.}
	\vspace{-0.15in}
	\label{fig:multiheadatt}
\end{figure}

Based on the above attention function in \eqref{dot-product}, \citet{vaswani2017attention} further develop a multi-head attention module, which allows the NMT model to jointly attend to information from different representations at different positions. In particular, we consider a multi-head attention module with $m$ heads. For the $i$-th head $H_i$, three point-wise linear transformations $W_{i,Q},~W_{i,K}~,~W_{i,V}\in\RR^{d\times d/m}$ are first applied to the input $Q$, $K$ and $V$, respectively, and then the scaled dot-product attention is applied: Let $\tilde{Q}_i=QW_{i,Q}$, $\tilde{K}_i=KW_{i,K}$ and $\tilde{V}=VW_{i,V}$,
\begin{align}\label{MHT}
H_i=\textrm{Attention}(\tilde{Q}_i,\tilde{K}_i,\tilde{V}_i).
\end{align}
Eventually, the final output applies a point-wise linear transformation $W_O\in\RR^{d\times d}$ to the concatenation of the output from all heads:
\begin{align*}
\textrm{MultiHead}(Q,K,V) = \textrm{Concat}(H_1,...,H_m)W_O.
\end{align*}
An illustrative example of the multihead attention architecture is provided in Figure~\ref{fig:multiheadatt}.

In addition to the above multi-head attention modules, each layer in the encoder and decoder in Transformer contains a point-wise two-layer fully connected feed-forward network.  

\vspace{-0.1in}
\section{Model} \label{sec:method}
\vspace{-0.1in}

We present our Transformer-based multi-domain neural machine translation model with word-level layer-wise domain mixing.

\vspace{-0.075in}
\subsection{Domain Proportion}\label{sec:domainprop}
\vspace{-0.075in}

Our proposed model is motivated by the observation that although every sentence in the training data has a domain label, a word in the sentence does not necessarily only belong to that single domain. Therefore, we assume that every word in the vocabulary has a domain proportion, which indicates its domain preference. Specifically, given the embedding $x\in\RR^{d}$ of a word, $k$ domains and $R\in\RR^{k\times d}$, our model represents the domain proportion by a smoothed softmax layer as follows, 
\begin{align*}
\cD(x) = (1-\epsilon)\cdot\textrm{softmax}(Rx)+\epsilon/k,
\end{align*}
where $\epsilon\in(0,1)$ is a smoothing parameter to prevent the output of $\cD(x)$ from collapsing towards $0$ or $1$. Specifically, setting $\epsilon$ as a large value encourages the word to be shared across domains.

\vspace{-0.075in}
\subsection{Word-Level Adaptive Domain Mixing}\label{sec:word_level_mixing}
\vspace{-0.075in}

\begin{figure}[!htb]
\centering
\vspace{-0.2in}
\includegraphics[width=0.3\textwidth]{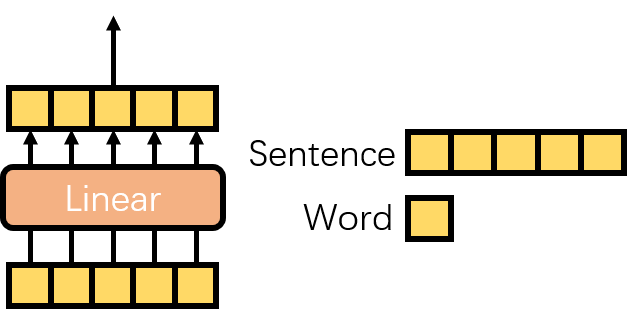}
\vspace{-0.1in}
\caption{The Point-wise Linear Transformations are applied at the word-level.}
\vspace{-0.15in}
\label{fig:pointwiselinear}
\end{figure}

\begin{figure}[!htb]
	\centering
	\vspace{-0.1in}
	\includegraphics[width=0.45\textwidth]{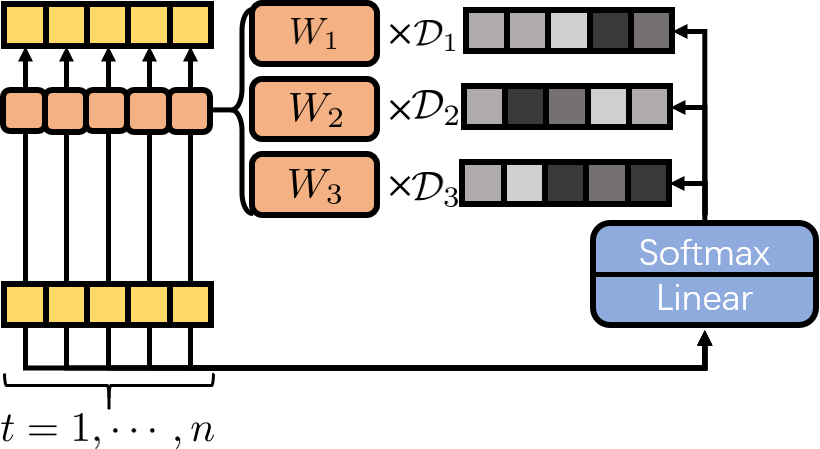}
	\vspace{-0.1in}
	\caption{Word-level mixing with 3 domains. For simplicity, we omit the subscripts $Q,i$.}
	\vspace{-0.1in}
	\label{fig:mixlinear}
\end{figure}

In our proposed model, each domain has its own multi-head attention modules. Recall that the point-wise linear transformations in the multi-head attention module $W_{i,Q}$'s, $W_{i,K}$'s, $W_{i,V}$'s and $W_O$ are applied to each word separately and identically, as shown in Figure~\ref{fig:pointwiselinear}.
Therefore, we can naturally integrate the domain proportions of the words with these multi-head attention modules. Specifically, we take the weighted averaging of the linear transformation based on the domain proportion $\cD(x)$. For example, we consider the point-wise linear transformations $\{W_{i,Q,j}\}_{j=1}^k$ on the $t$-th word of the input, $Q_t$, of all domains. The mixed linear transformation can be written as 
\[
\textstyle\overline{Q}_{i,t}=\sum_{j=1}^k Q_t^\top W_{i,Q,j} \cD_{Q,j}(Q_t),
\]
where $\cD_{Q,j}(Q_t)$ denotes the $j$-th entry of $\cD_Q(Q_t)$, and $\cD_Q$ is the domain proportion layer related to $Q$. Then we only need to replace $\tilde{Q}_i$ in \eqref{MHT} with
\begin{align*}[\overline{Q}_{i,1},...,\overline{Q}_{i,n}].
\end{align*}
An illustrative example is presented in Figure~\ref{fig:mixlinear}. For other linear transformations, we applied the domain mixing scheme in the same way.
We remark that the Transformer model, though does not have any explicit recurrent structure, handles the sequence through adding additional positional embedding for each word (in conjunction with sequential masking). Therefore, if a word appears in different positions of a sentence, its corresponding embedding is different. This indicates that the domain proportions of the same word can also be different across positions. This feature makes our model more flexible, as the same word in different positions can carry different domain information.


%

\vspace{-0.075in}
\subsection{Layer-wise Domain Mixing}\label{sec:layerwise}
\vspace{-0.075in}

\begin{figure}[!htb]
	\centering
	\vspace{-0.25in}
	\includegraphics[width=0.45\textwidth]{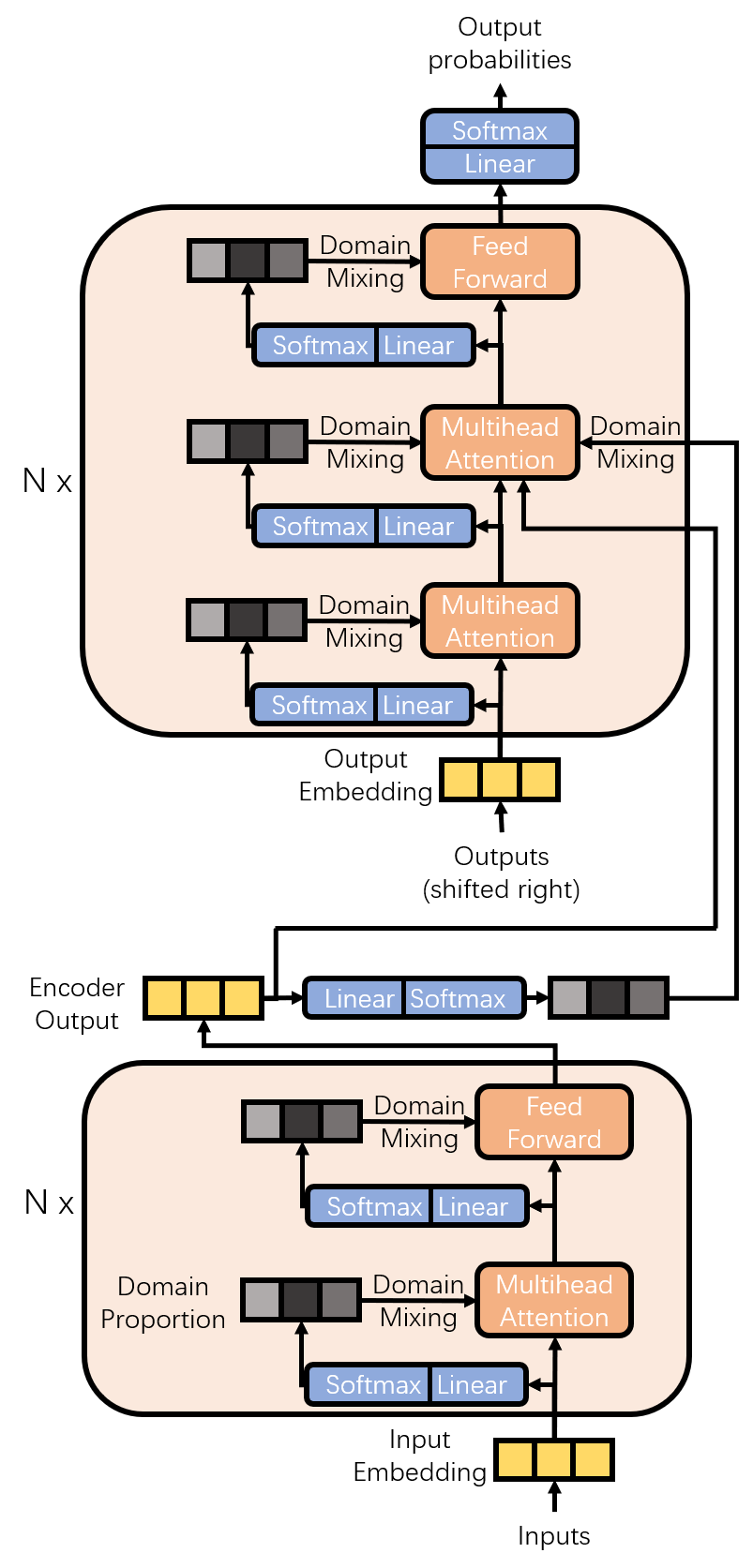}
	\vspace{-0.15in}
	\caption{Illustration of Our Multi-domain NMT Model: Normalization and residual connection are omitted for simplicity. For all other detail, please refer to \citet{vaswani2017attention}. }
	\vspace{-0.2in}
	\label{fig:fullmodel}
\end{figure}

Recall that the Transformer model contains multiple multi-head attention modules/layers. Therefore, our proposed model inherits the same architecture and applies the word-level domain mixing to all these attention layers. Since the words have different representations at each layer, the corresponding domain proportions at each layer are also different, as shown in Figure \ref{fig:fullmodel}. In addition to the multi-head attention layers, we also apply similar word-level domain mixing to the point-wise two-layer fully connected feed-forward network.

The layer-wise domain mixing allows the domain proportions to be context dependent. This is because the domain proportions are determined by the word embedding, and the word embedding at top layers is essentially learnt from the representations of all words at bottom layers. As a result, when the embedding of a word at some attention layer is already learned well through previous layers (in the sense that it contains sufficient contextual information and domain knowledge), we no longer need to borrow knowledge from other domains to learn the embedding of the word at the current layer. Accordingly, the associated domain proportion is expected to be skewed and discourages knowledge sharing across domains. This makes the process of knowledge sharing of our model more adaptive.




\vspace{-0.05in}
\subsection{Training} \label{sec:training}
\vspace{-0.05in}

Recall that $\cH$ denotes the encoder, $\cF$ denotes the decoder, and $\cD$ denotes the domain proportion. Define $\Theta=\{\cF,\cH,\cD\}$. The proposed model can be efficiently trained by minimizing a composite loss function defined as follows,
\begin{align*}
L^* = L_{\rm gen}(\Theta) + L_{\rm mix}(\Theta),
\end{align*}
where $L_{\rm gen}(\Theta)$ denotes the cross-entropy loss over the training data $\{\xb_i, \yb_i\}_{i=1}^n$, and $L_{\rm mix}(\Theta)$ denotes the cross entropy loss over the words/domain (hard) labels.

For $L_{\rm mix}(\Theta)$, the domain labels are obtained from the training data. Specifically, for all words in a sentence belonging to the $J$-th domain, we specify their domain hard labels as $J$. Then given the embedding $x$ of a word, we compute the cross entropy loss of its domain proportion $\cD(x)$ as $-\log(\cD_J(x))$. Accordingly, $L_{\rm mix}(\Theta)$ is the sum of the cross entropy loss over all such pairs of word/domain label of the training data.


\vspace{-0.1in}
\section{Experiment} \label{sec:exp}
\vspace{-0.1in}

We conduct experiments on three different machine translation tasks: 
\smallskip

\noindent$\bullet$ \textbf{English-to-German}. We use a dataset from two domains: News and TED. We collect the News domain data from Europarl \citep{koehn2005europarl} and the TED domain data from IWLST \citep{cettolo2014report}.

\smallskip

\noindent$\bullet$ \textbf{English-to-French} We use a dataset containing two domains: TED and Medical domain. We collect TED domain data from IWLST \citep{cettolo2017overview} and medical domain data from Medline \citep{yepes2017findings}. 

\smallskip

\noindent$\bullet$ \textbf{Chinese-to-English} We use a dataset containing four domains: News, Speech, Thesis and Laws. We collect the Laws, Speech, and Thesis data from UM-Corpus \citep{tian2014corpus}, and the News data from LDC \citep{openldc}. The translation from Chinese-to-English is inherently difficult. The four-domains setting makes it even more challenging. This dataset is also used in \citet{zeng2018multi}.

\smallskip

The sizes of training, validation, and testing sets for different language pairs are summarized in Table~\ref{tab:data}. We tokenize English, German and French sentences using MOSES script \citep{koehn2007moses} and perform word segmentation on Chinese sentences using Stanford Segmenter \citep{tseng2005conditional}. All sentences are then encoded using byte-pair encoding \cite{sennrich2015neural}. We evaluate the performance using two metrics: BLEU \cite{papineni2002bleu} and perplexity following the default setting in fairseq with beam search steps of $5$.

\begin{table}[!hbt]
	\centering
	\small
	\vspace{-0.05in}
	\begin{tabular}{c|cccc}
	    \hline
		\textbf{Language} & \textbf{Domain} & \textbf{Train} & \textbf{Valid} & \textbf{Test}  \\
		\hline
		\hline
		\multirow{2}{*}{\textbf{EN-DE}} 
		& News & 184K & 18K & 19K  \\
		& TED & 160K & 7K & 7K  \\
		\hline
		\multirow{2}{*}{\textbf{EN-FR}} 
		& TED & 226K & 10K & 10K  \\
		& MEDICAL & 516K & 25K & 25K  \\
		\hline
		\multirow{4}{*}{\textbf{ZH-EN}} 
		& Laws & 219K & 600 & 456  \\
		& News & 300K & 800 & 650  \\
		& Speech & 219K & 600 & 455  \\
		& Thesis & 299K & 800 & 625  \\
	    \hline
	\end{tabular}
	\vspace{-0.1in}
	\caption{The numbers of sentences in the datasets.}
	\vspace{-0.15in}
	\label{tab:data}
\end{table}

\vspace{-0.1in}
\subsection{Baselines} 
\vspace{-0.1in}

Our baselines include the Transformer models trained using data from single and all domains. We also include several domain aware embedding based methods, which train the embedding of the encoder along with domain information. 

\smallskip

\noindent $\bullet$\textbf{Multitask Learning (MTL)} proposed in \citet{britz2017effective} uses one sentence-level domain classifier to train the embedding. Note that their classifier is only used to predict the domain, while our model uses multiple word-level domain classifiers to obtain the domain proportions for different layers (further used for domain mixing). 

\smallskip

\noindent $\bullet$ \textbf{Adversarial Learning (AdvL)} proposed in \citet{britz2017effective} is a variant of MTL, which flips the gradient before it is back-propagated into the embedding. This encourages the embedding from different domains to be similar.

\smallskip

\noindent $\bullet$ \textbf{Partial Adversarial Learning (PAdvL)} To combine the advantages of the above two methods, we split the embedding into half of multitask part and half of adversarial part. 

\smallskip

\noindent $\bullet$ \textbf{Word-Level Domain Context Discrimination (WDC)} \citet{zeng2018multi} integrates MTL and AdvL with word-level domain contexts. This method requires the dimension of the embedding to be doubled and, thus, is not directly applicable in Transformer. We use a point-wise linear transformation to reduce the dimension.

Moreover, \citet{zeng2018multi} consider the word-level domain aware weighted loss (\textbf{WL}). Specifically, they assign a domain-aware attention weight $\beta_j$ to the $j$-th position in the output sentence, and the corresponding weighted loss is:
\[
\textstyle L_{\textrm{gen}} = -\frac{1}{n} \sum_{j=1}^n (1+\beta_j) \log p(y_j | \xb, y_{<j}). \]
Here $\beta_j$ is obtained by an attention based domain classifier built upon the last hidden layer. 

\vspace{-0.075in}
\subsection{Details of Our Implementation} \label{sec:implementation_details}
\vspace{-0.075in}

All of our experiments are conducted under fairseq \citep{ott2019fairseq} environment. We follow the fairseq re-implementation of Transformer designed for IWLST data. Specifically, the re-implementation of Transformer differs from \citet{vaswani2017attention} in that the embedding dimension is $512$ for both the encoder and decoder, the number of heads is $6$, and the embedding dimension in the feed-forward layer is $1024$. In terms of the optimization, we follow the training recipe provided by fairseq. Specifically, we use Adam optimizer \citep{kingma2014adam} with $\beta_1=0.9, \beta_2=0.98$ with a weight decay parameter of $10^{-4}$. The learning rate follows the inverse square root schedule \cite{vaswani2017attention} with warm-up steps of $4000$, initial warm-up learning rate of $10^{-7}$, and the highest learning rate of $5 \times 10^{-4}$. For effective training, $L_{\textrm{gen}}$ is replaced by a label-smoothing cross-entropy loss with a smoothing parameter of $0.1$ \cite{szegedy2016rethinking}. 

\begin{figure}[!htb]
	\centering
	\vspace{-0.15in}
	\includegraphics[width=0.3\textwidth]{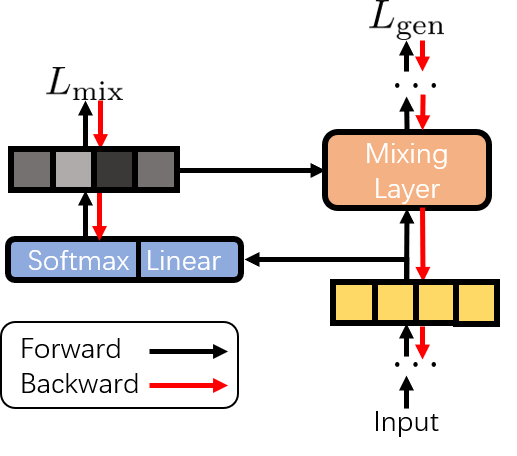}
	\vspace{-0.15in}
	\caption{Computational graph for training the domain proportion layers.}
	\vspace{-0.15in}
	\label{fig:detached_train}
\end{figure}

For our domain mixing methods, we set the smoothing parameter $\epsilon$ of the domain proportion as $0.05$. Besides applying domain mixing to both the encoder and decoder (\textbf{E/DC}), we consider applying domain mixing to only the \textbf{Encoder}. The domain proportion layers $\cD$ are only used for estimating the domain proportion and should not intervene in the training of the translation model. So the gradient propagation is cut off between the Transformer and the domain proportion as Figure~\ref{fig:detached_train} shows. More discussion about the training procedure can be found in Section~\ref{sec:grad_flow}.

\vspace{-0.075in}
\subsection{Experimental Results} 
\vspace{-0.075in}

Table~\ref{tab:en2deExp} shows the BLEU scores of the baselines and our domain mixing methods for English-to-German translation. As can be seen, our methods outperform the baselines on both domains. 

\begin{table}[!hbt]
\vspace{-0.1in}
    \centering
    \begin{tabular}{c|c|c}
    \hline
    Method & News & TED  \\
    \hline
    \hline
    \multicolumn{3}{c}{Direct Training}\\
    \hline
    News & 26.09& 6.15\\
    TED & 4.90& 29.09\\
    News + TED & 26.06& 28.11\\
    \hline
    \multicolumn{3}{c}{Embedding based Methods}\\
    \hline
    MTL & 26.90& 29.27\\
    AdvL  & 25.68& 27.46\\
    PAdvL& 27.06& 29.49\\
    WDC + WL& 27.25 & 29.43\\
    \hline
    \multicolumn{3}{c}{Our Domain Mixing Methods}\\
    \hline
    Encoder & \textbf{27.78}& 30.30\\
    Encoder + WL & 27.67& 30.11\\
    E/DC & 27.58 & \textbf{30.33} \\
    E/DC + WL& 27.55& 30.22\\
    \hline
    \end{tabular}
     \vspace{-0.05in}
    \caption{English-to-German.}
     \vspace{-0.25in}
    \label{tab:en2deExp}
\end{table}

\begin{figure}[!htb]
    \centering
    \vspace{-0.075in}
    \includegraphics[width=0.6\textwidth]{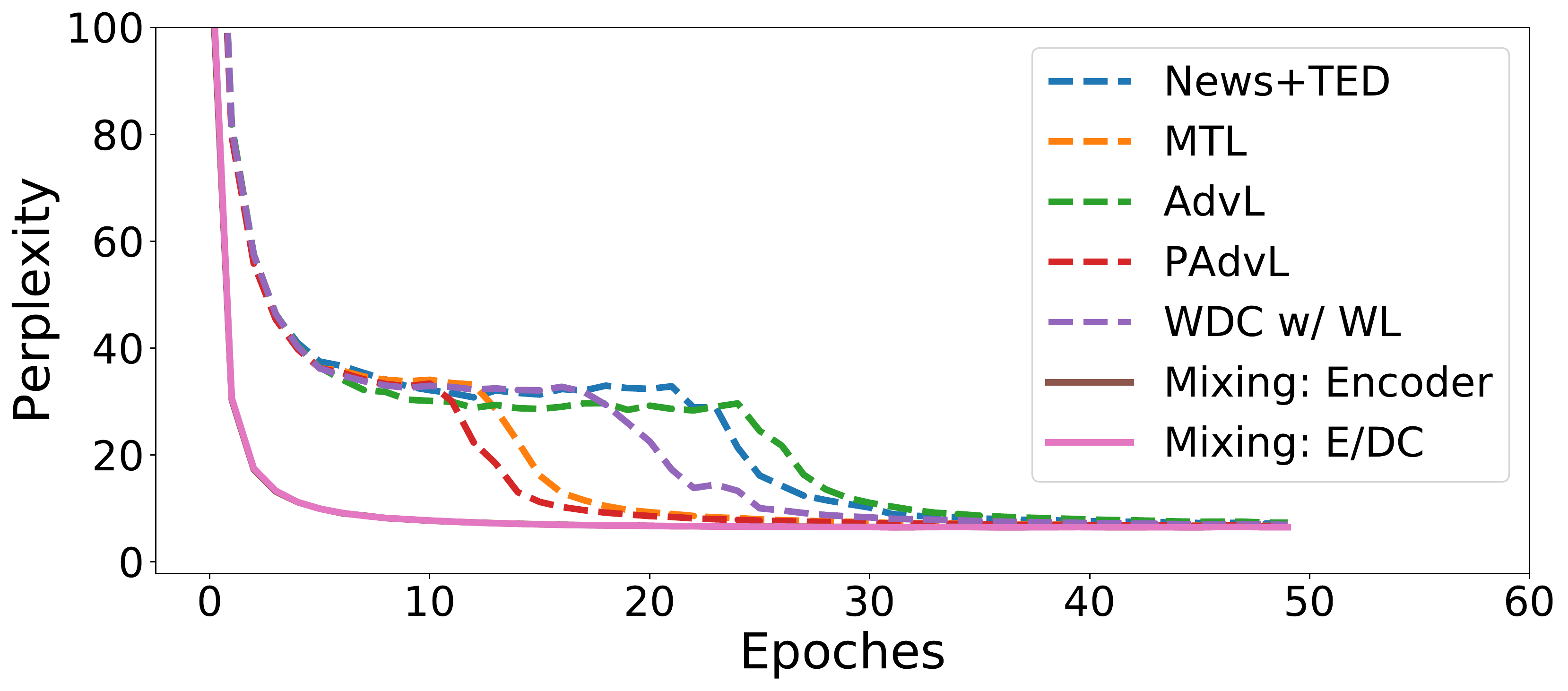}
    \vspace{-0.15in}
    \caption{Perplexity v.s. Number of epochs for English-to-German.}
    \label{fig:en2dePPL}
    \vspace{-0.125in}
\end{figure}

We also compare the perplexity on the validation set in Figure~\ref{fig:en2dePPL}. As can be seen, our domain mixing methods converge faster than the baselines and all methods converge after 50 epochs. We also observe that the baselines get stuck at plateaus at the early stage of training. The possible reason is that their training enforces one unified model to fit data from two different domains simultaneously, which is computationally more difficult. 

Table~\ref{tab:en2frExp} shows the BLEU scores of the baselines and our domain mixing methods for English-to-French translation. Note that though the data from the Medical and TED domains are slightly imbalanced (about 1:2.5), our methods can still outperform the baselines on both domains.


\begin{table}[!hbt]
    \centering
    \vspace{-0.05in}
    \begin{tabular}{c|c|c}
    \hline
    Method & TED & Medical  \\
    \hline
    \hline
    \multicolumn{3}{c}{Direct Training}\\
    \hline
    TED & 28.22& 7.32\\
    Medical & 7.03& 53.73\\
    Medical + TED & 39.21& 53.40\\
    \hline
    \multicolumn{3}{c}{Embedding based Methods}\\
    \hline
    MTL & 39.14& 53.37\\
    AdvL  & 39.54& 53.46\\
    PAdvL& 39.56& 53.23\\
    WDC + WL&39.79 &53.85  \\
    \hline
    \multicolumn{3}{c}{Our Domain Mixing Methods}\\
    \hline
    Encoder & 40.30& 54.05\\
    Encoder + WL & 40.43& 54.14\\
    E/DC & 40.52&	54.28 \\
    E/DC + WL & \textbf{40.60} &	\textbf{54.39} \\
	    \hline
    \end{tabular}
    
    \vspace{-0.1in}
    \caption{English-to-French.}
    \vspace{-0.15in}
    \label{tab:en2frExp}
\end{table}

\begin{table}[!hbt]
    \vspace{-0.05in}
	\centering
	\begin{tabular}{c|c|c|c|c}
	    \hline
		Method & Laws & News & Speech & Thesis   \\
		\hline
		\hline
		\multicolumn{5}{c}{Direct Training}\\
		\hline
		Laws &  51.98&	3.80&	2.38&	2.64\\
		News &  6.88&	31.99&	8.12&	4.17\\
		Speech &  3.33&	4.90&	18.63&	3.08\\
		Thesis &  5.90&	5.55&	4.77&	11.06\\
		Mixed &  48.87&	26.92&	16.38&	12.09\\
		\hline
		\multicolumn{5}{c}{Embedding based Methods}\\
		\hline
		MTL & 49.14&	27.15&	16.34&	11.80\\
		AdvL  & 48.93&	26.51&	16.18&	12.08\\
		PAdvL & 48.72&	27.07&	15.93&	\textbf{12.23}\\
		WDC + WL& 42.16& 25.81& 15.29& 10.14\\
		\hline
		\multicolumn{5}{c}{Our Domain Mixing Methods}\\
		\hline
		Encoder & 50.21&	27.94&	16.85&	12.03\\
		Encoder + WL & 50.11& 27.48&	16.79& 11.93\\
		E/DC &\textbf{50.64}&	\textbf{28.48}&	17.41&	11.71  \\
		E/DC + WL & 50.04& 28.17&  \textbf{17.60}& 11.59\\
	    \hline 
	\end{tabular}
	\caption{Chinese-to-English.} 
	\label{tab:zh2enExp_stable}
\end{table}

Table~\ref{tab:zh2enExp} shows the BLEU scores of the baselines and our domain mixing methods for Chinese-to-English translation. As can be seen, our methods outperform the baselines on all domains except Thesis. We remark that the translation for the Thesis domain is actually very difficult, and all methods obtain poor performance.

Moreover, we find that for Chinese-to-English task, all our baselines are sensitive to the architecture of the Transformer. Their training will fail, if we place the layer normalization at the end of each encoder and decoder layer (as \citet{vaswani2017attention} suggest). Therefore, we move the layer normalization to their beginnings. Surprisingly, our domain mixing methods are very stable regardless of the position of the layer normalization. More details can be found in Table~\ref{tab:zh2enExp} of Appendix~\ref{sec:extraexp}.

\vspace{-0.075in}
\subsection{Ablation Study}
\vspace{-0.075in}

We further shows that the performance gains are from the domain mixing methods, instead of from the increased model capacity. Table~\ref{tab:ablationExp} shows the BLEU scores with and w/o using domain labels under equal model size. We keep the model size consistent by adopting the same network structure and the same number of parameters as in the domain mixing methods. The only difference is that we remove domain label to guide the training of domain proportion, i.e., only $L_{\rm gen}$ is used in the training loss, and $L_{\rm mix}$ is removed. Training w/o domain labels shows a slight improvement over baseline, but is still significantly worse than our proposed method for most of the tasks. Therefore, we can conclude that our proposed domain mixing approach indeed improves performance.

\begin{table}[!hbt]
	\vspace{-0.05in}
	\centering
	\begin{tabular}{c|c|c|c}
		\hline
		Method & Direct Training & w/o DL & with DL (Ours)  \\
		\hline
		\hline
		\multicolumn{4}{c}{English-to-Germany}\\
		\hline
		News &  26.06 &	26.25 &	\textbf{27.78}\\
		TED & 28.11 & 28.27 & \textbf{30.30} \\
		\hline
		\multicolumn{4}{c}{English-to-French}\\
		\hline
		TED & 39.21 & 39.39 & \textbf{40.30} \\
		Medical & 53.40 & 53.33 & \textbf{54.05} \\
		\hline
		\multicolumn{4}{c}{Chinese-to-English}\\
		\hline
		Laws & 48.87 & 48.96 & \textbf{50.21} \\
		News & 26.92 & 27.02 & \textbf{27.94} \\
		Speech & 16.38 & 16.15 & \textbf{16.85} \\
		Thesis & \textbf{12.09} & 12.03 & 12.03 \\
		\hline 
	\end{tabular}
	\caption{BLEU Scores with and w/o domain labels (DL) under equal model capacity.} 
	\label{tab:ablationExp}
\end{table}
\vspace{0.1in}

\vspace{-0.075in}
\subsection{Visualizing Domain Proportions} \label{sec:exp_domainprop}
\vspace{-0.075in}

To further investigate our domain mixing methods, we plot the domain proportions of the word embedding at different layers. A uniform proportion, e.g., $(0.5,0.5)$, is encouraging knowledge sharing across domains, while a skewed proportion, e.g., $(0.1,0.9)$, means there is little knowledge to share across domains.
\begin{figure}[htb!]
	\centering
	\includegraphics[width=0.7\textwidth]{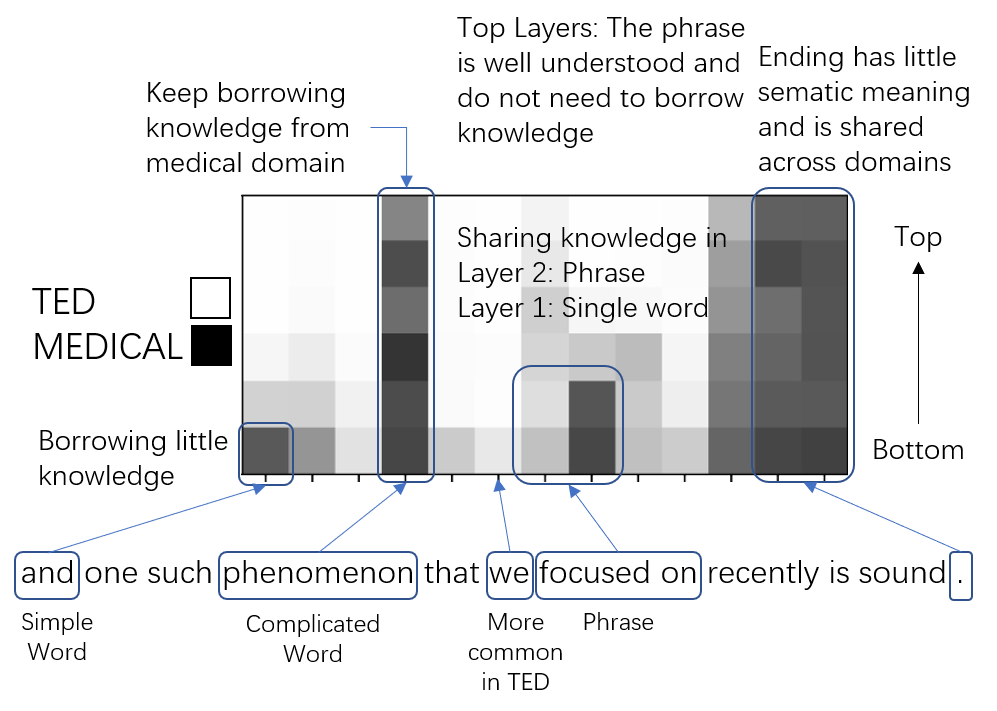}
	\caption{Domain proportion of a sentence from the TED domain for English-to-French task. The domain proportion is extracted from all layers of the encoder.} 
	\label{fig:explain}
\end{figure}
Figure~\ref{fig:explain} illustrates how the knowledge sharing is controlled via the domain proportion. The selected sentence is from the English-to-French task, containing TED and Medical domains. Specifically, we observe :

\smallskip

\noindent $\bullet$ The domain proportions of different words at different layers have various patterns.

\smallskip

\noindent $\bullet$ At the bottom layers, the domain proportion of a word is closely related to its frequency of occurrence.

\smallskip

\noindent $\bullet$ Some words with simple semantic meanings do not need to borrow much knowledge from other domains, e.g., \textit{and}; Some other words need to borrow knowledge from other domains to better understand their own semantic meaning. For example, the word \textit{phenomenon} keeps borrowing/sharing knowledge from/to the medical domain at every layer.

\smallskip

\noindent $\bullet$ The ending of the sentence only conveys a stopping signal, and thus is shared across all domains.

\smallskip

\begin{figure}[htb!]
	\centering
	\includegraphics[width=0.7\textwidth]{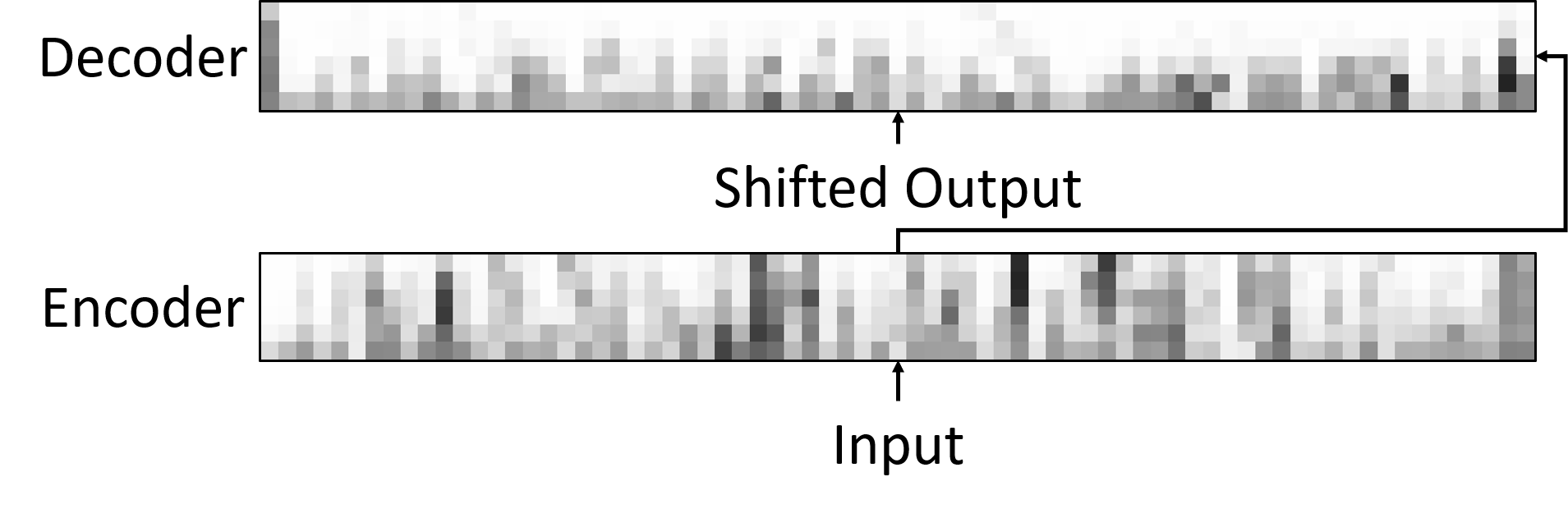}
	\caption{Domain proportions of a sentence pair for English-to-German task. White represents the News domain and black represents the TED domain. The domain proportions of both the encoder (bottom) and the decoder (top) are presented. } 
	\label{fig:domainprop_senence}
\end{figure}

\noindent $\bullet$ The domain proportions at the bottom layers tend to be more diverse, while those at the top layers tend to be more skewed, as shown in Figure~\ref{fig:domainprop_senence} for English-to-German task.

\smallskip

\begin{figure}[htb!]
\centering
	\begin{tabular}{cccccc}
		Layer-1 &2&3&4&5&6\\
		\hline
		\includegraphics[width=0.13\textwidth]{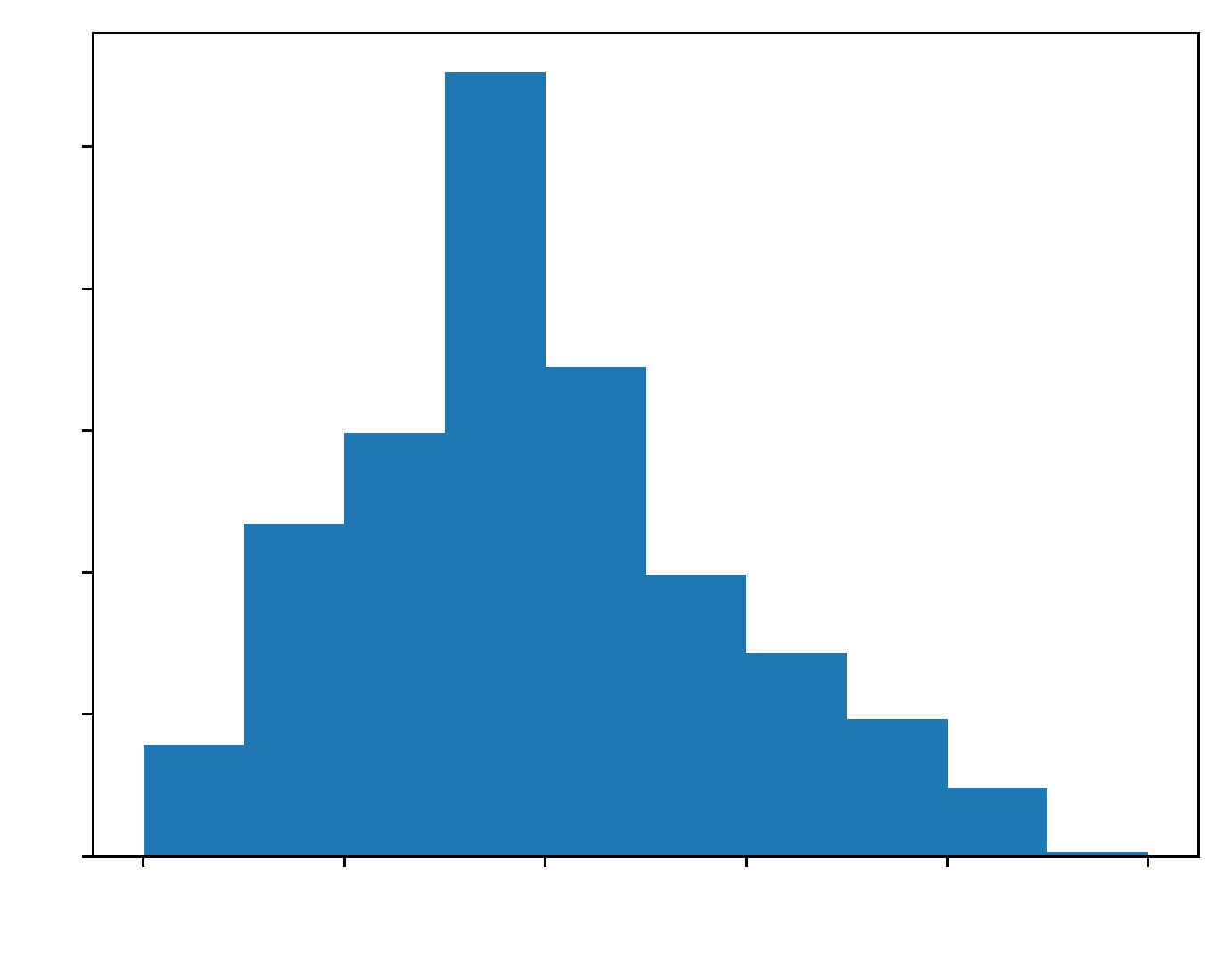}&
		\includegraphics[width=0.13\textwidth]{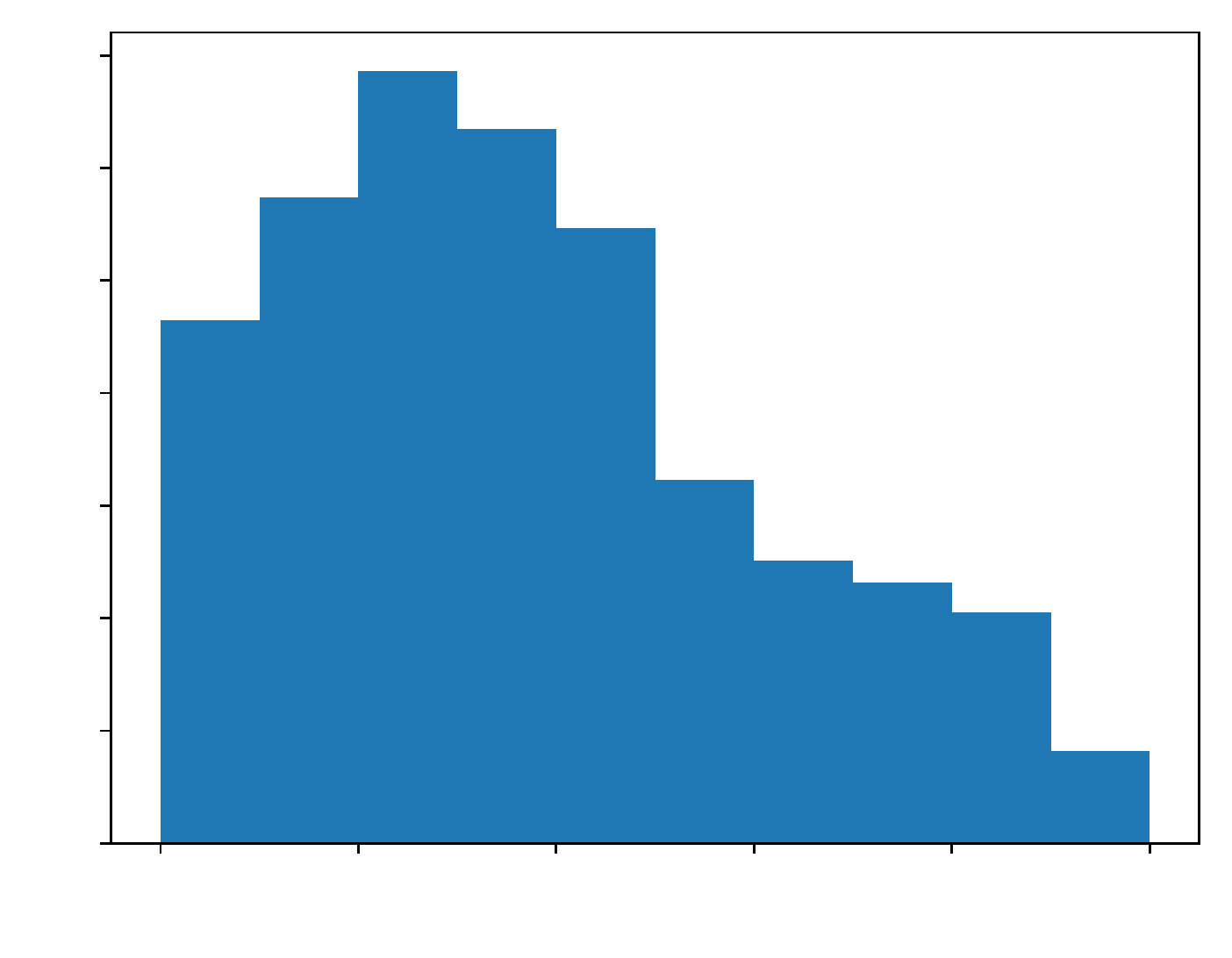}&
		\includegraphics[width=0.13\textwidth]{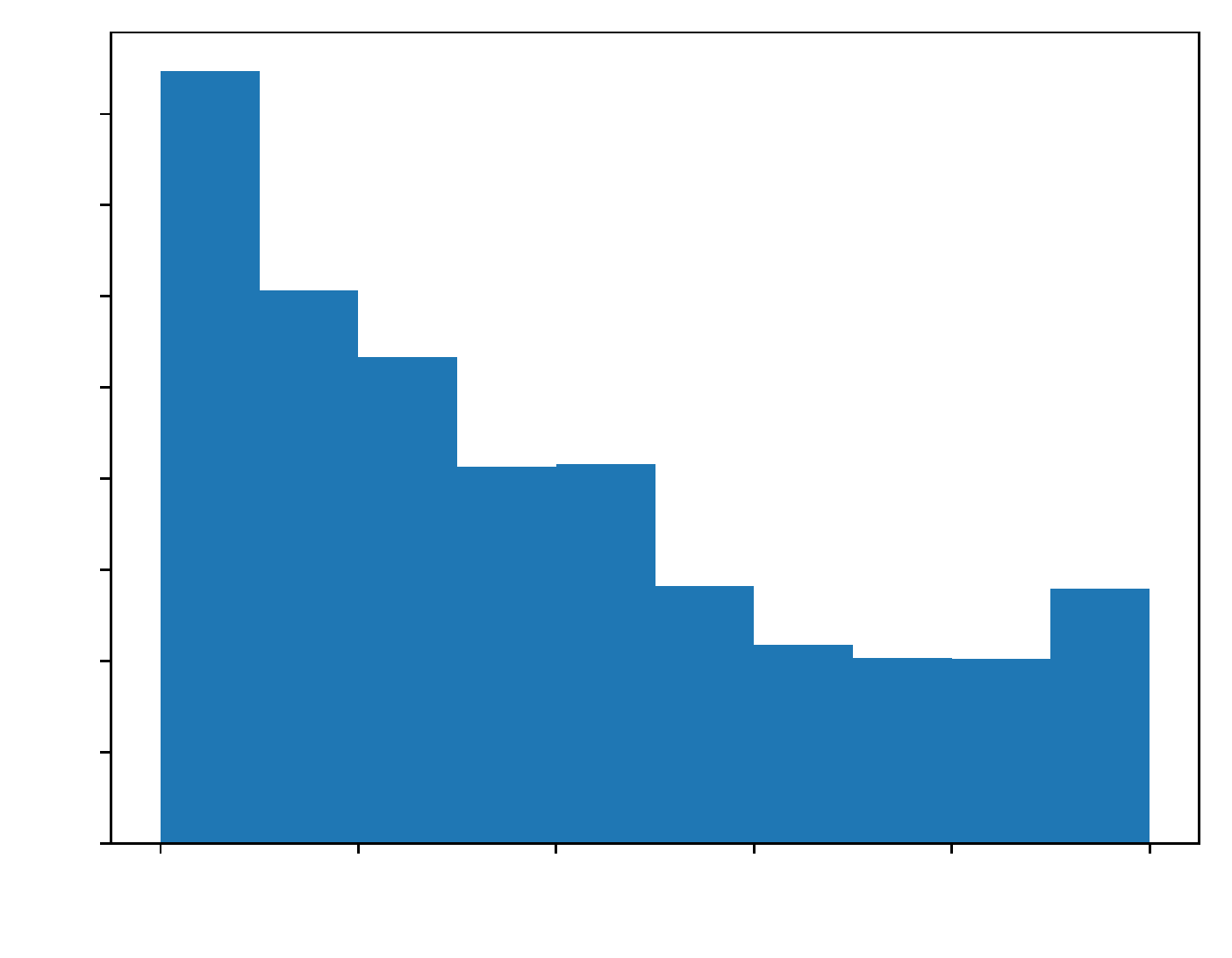}&
		\includegraphics[width=0.13\textwidth]{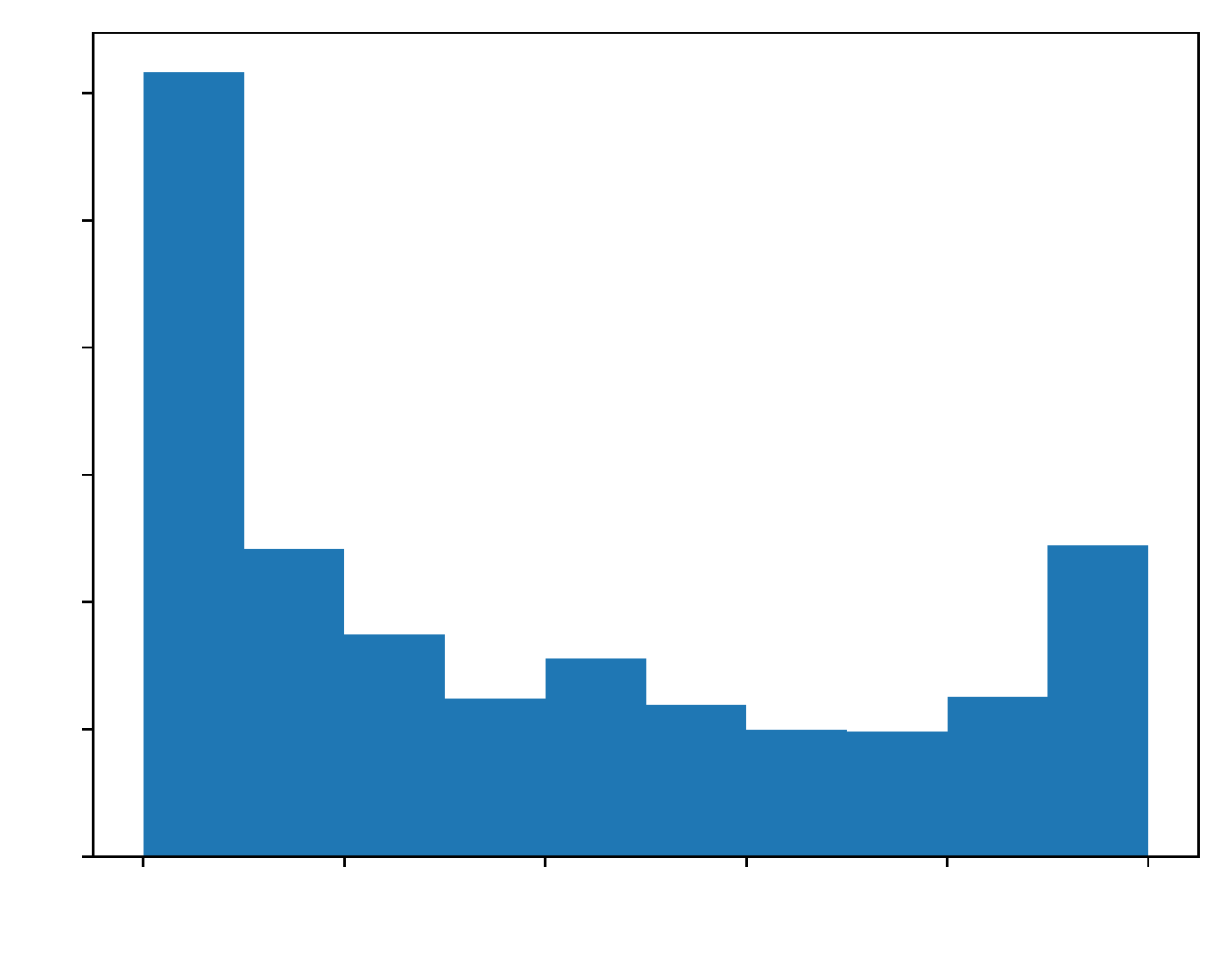}&
		\includegraphics[width=0.13\textwidth]{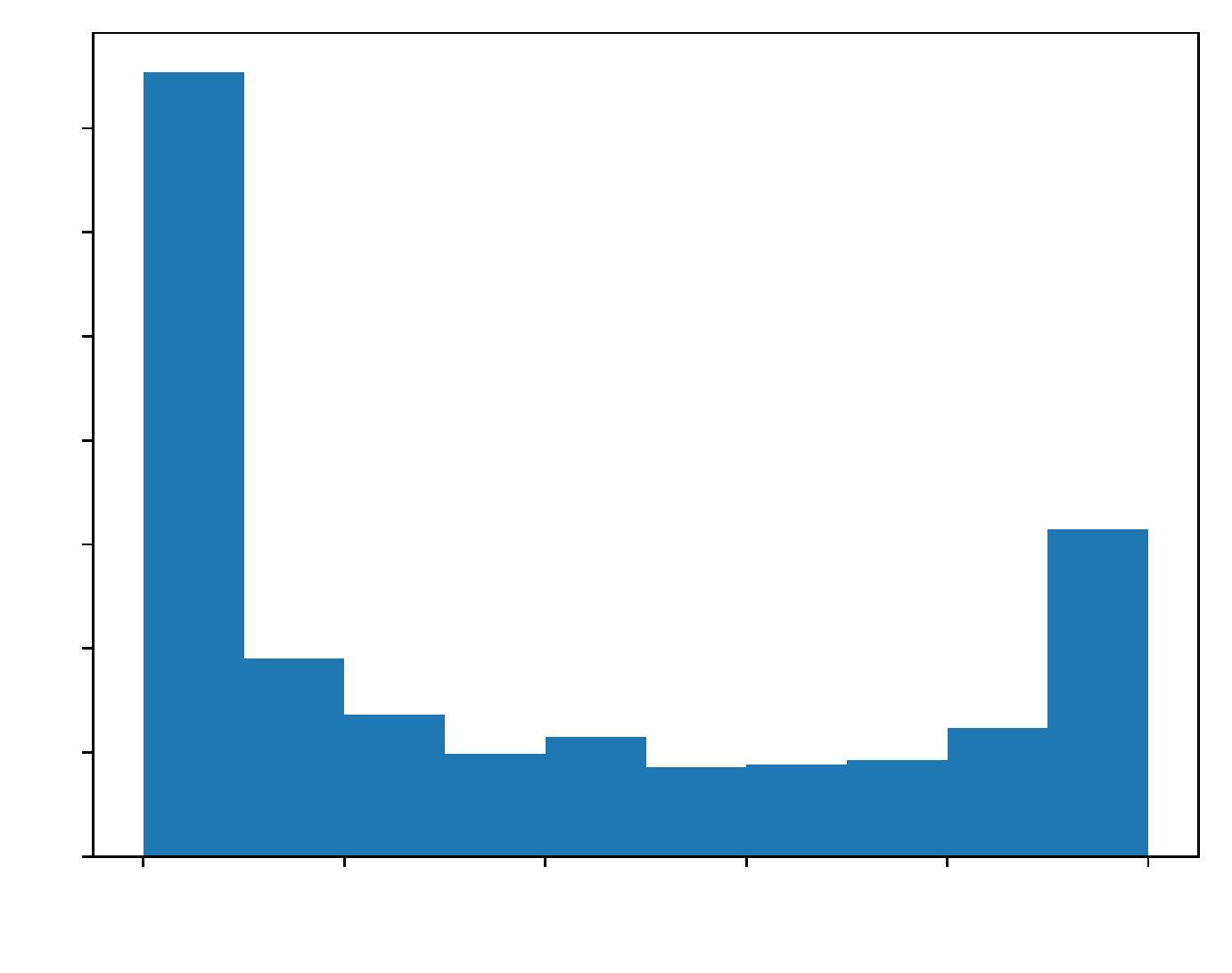}&
		\includegraphics[width=0.13\textwidth]{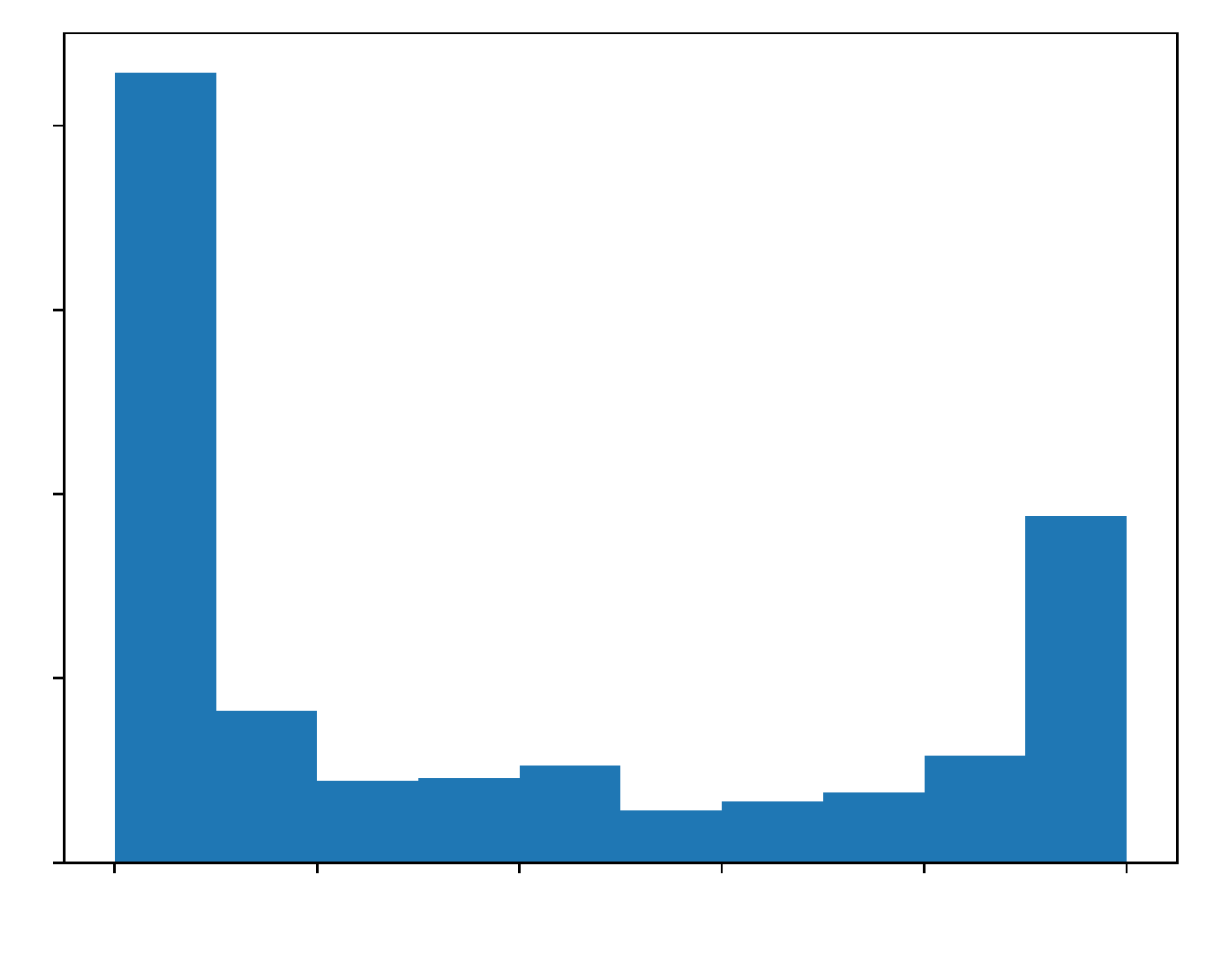} \vspace{-0.1in}\\
		\multicolumn{6}{c}{Encoder} \\
		\includegraphics[width=0.13\textwidth]{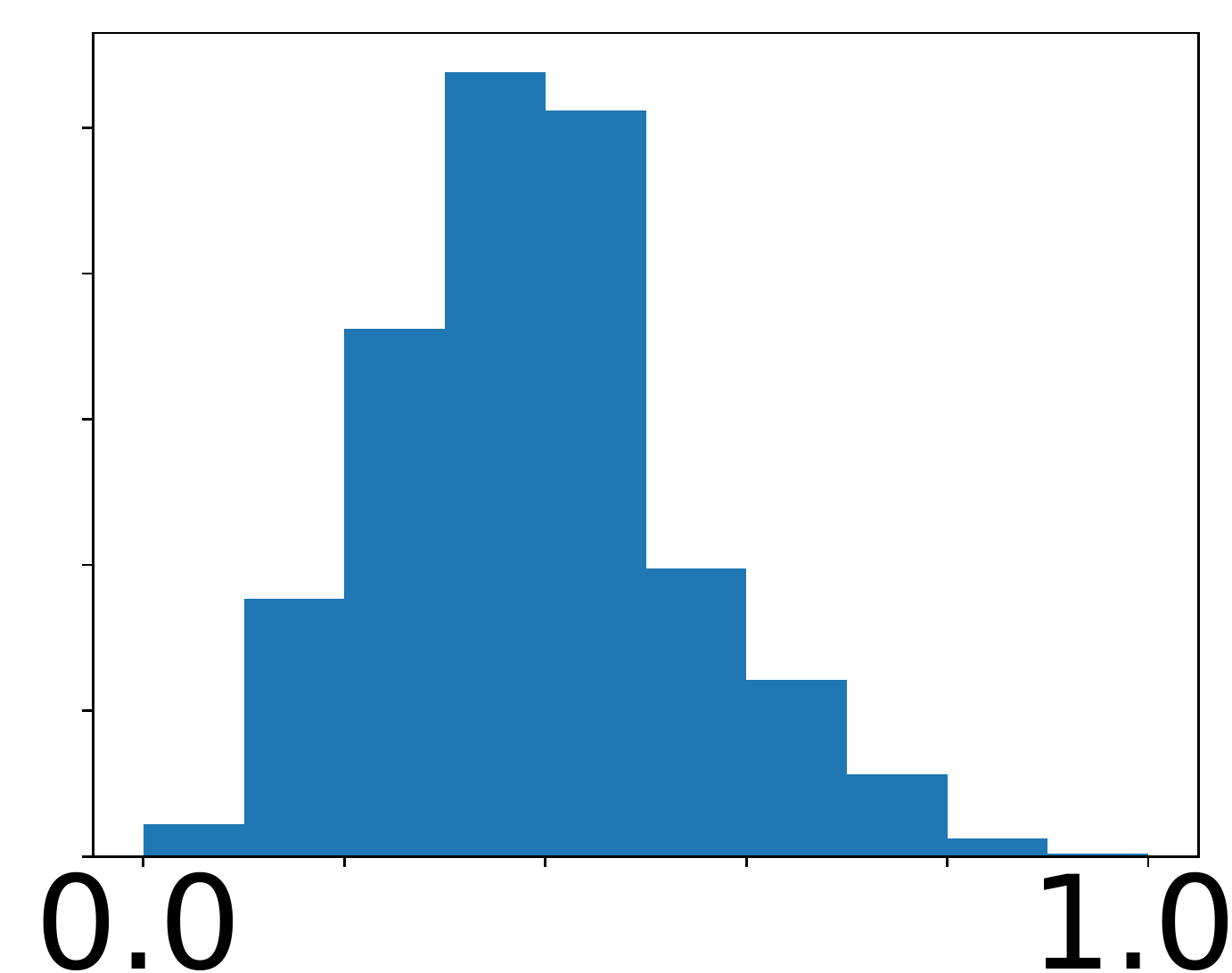}&
		\includegraphics[width=0.13\textwidth]{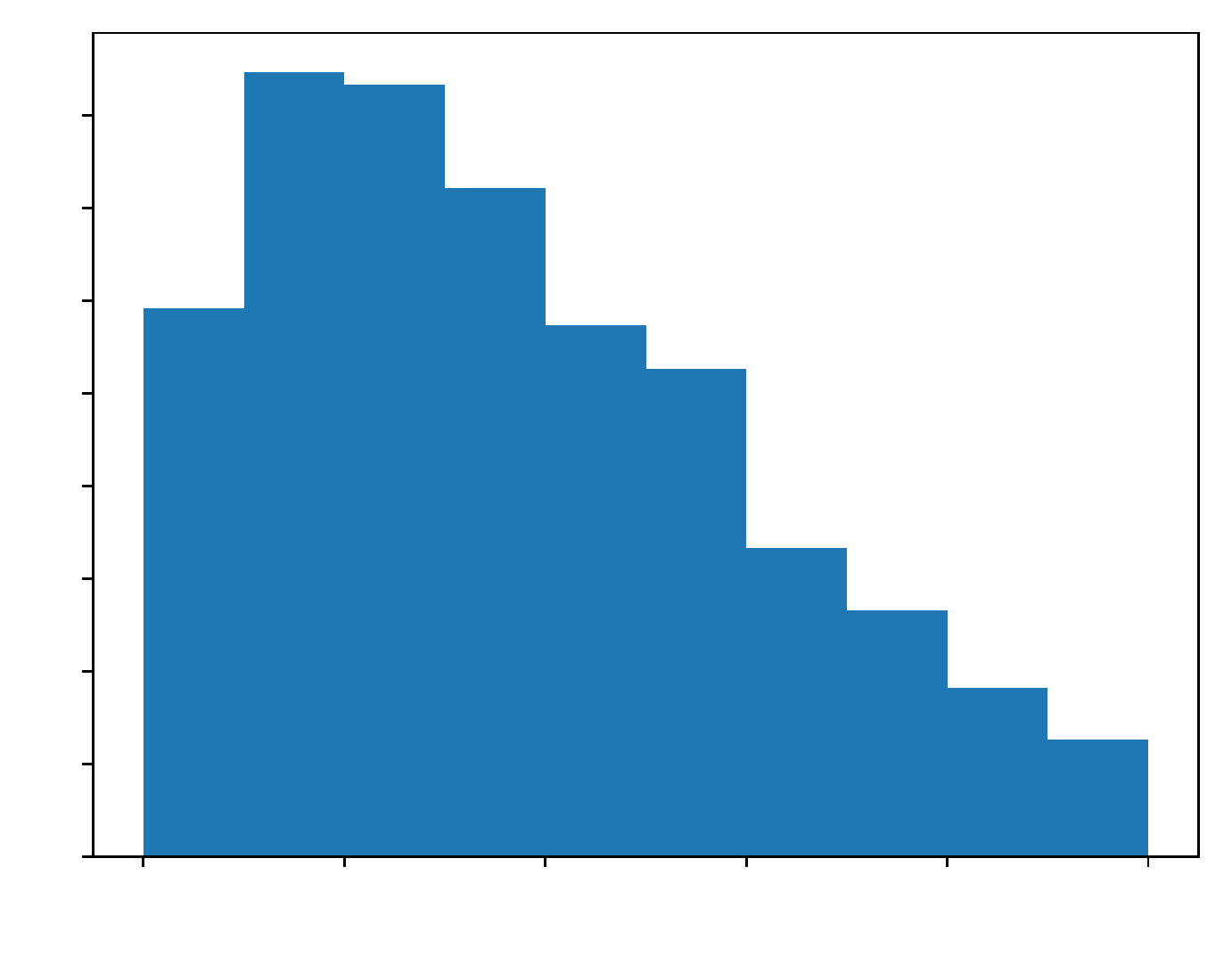}&
		\includegraphics[width=0.13\textwidth]{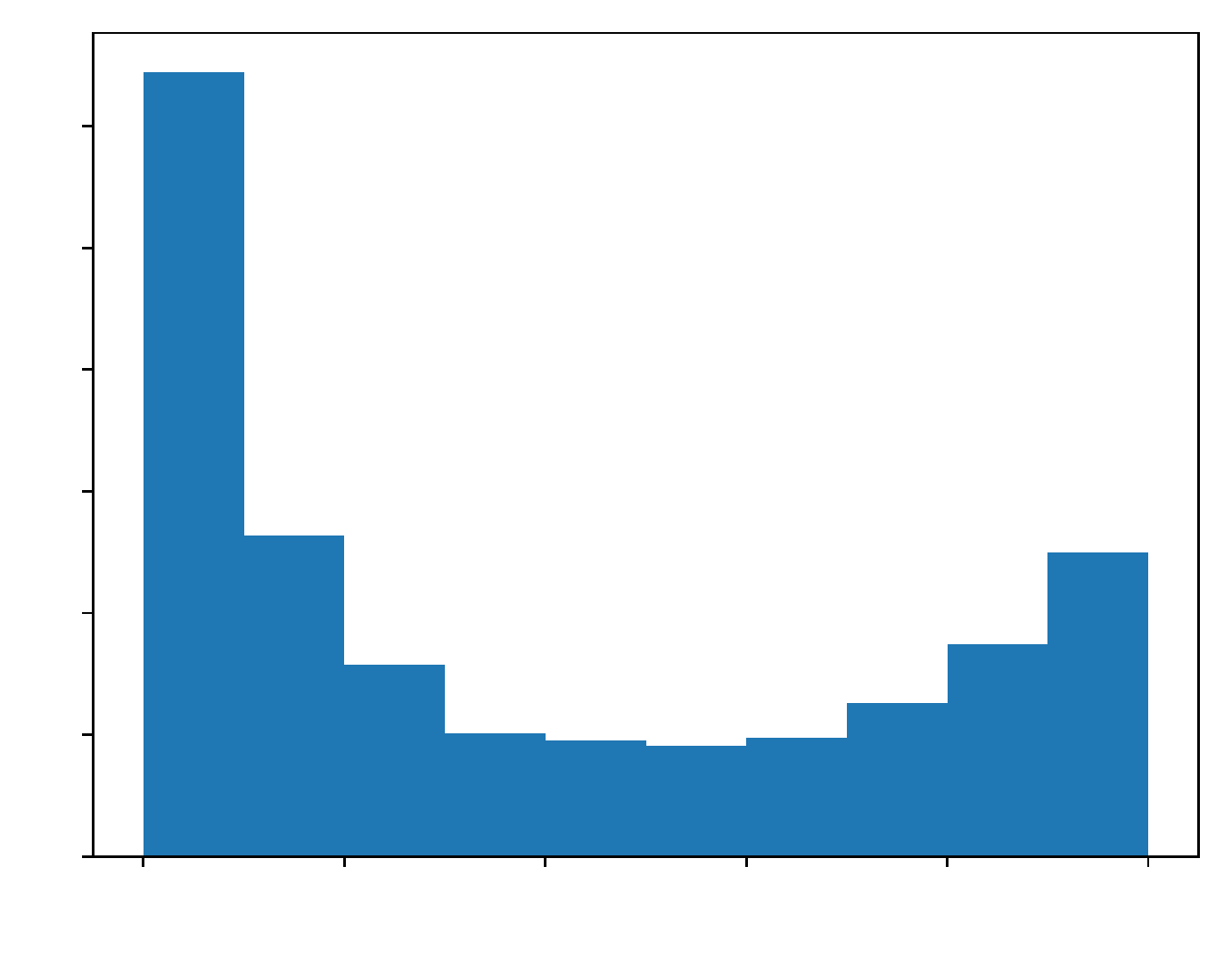}&
		\includegraphics[width=0.13\textwidth]{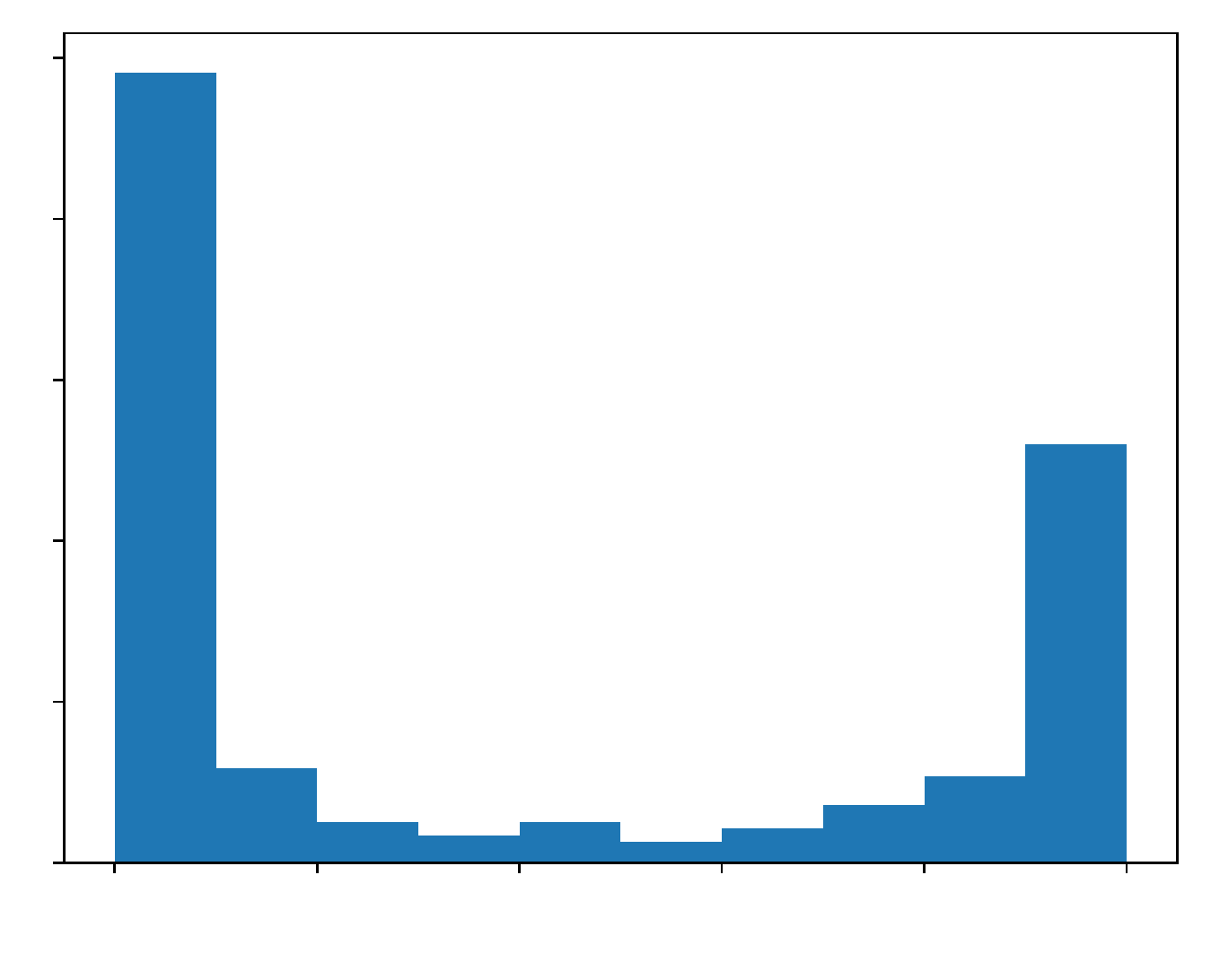}&
		\includegraphics[width=0.13\textwidth]{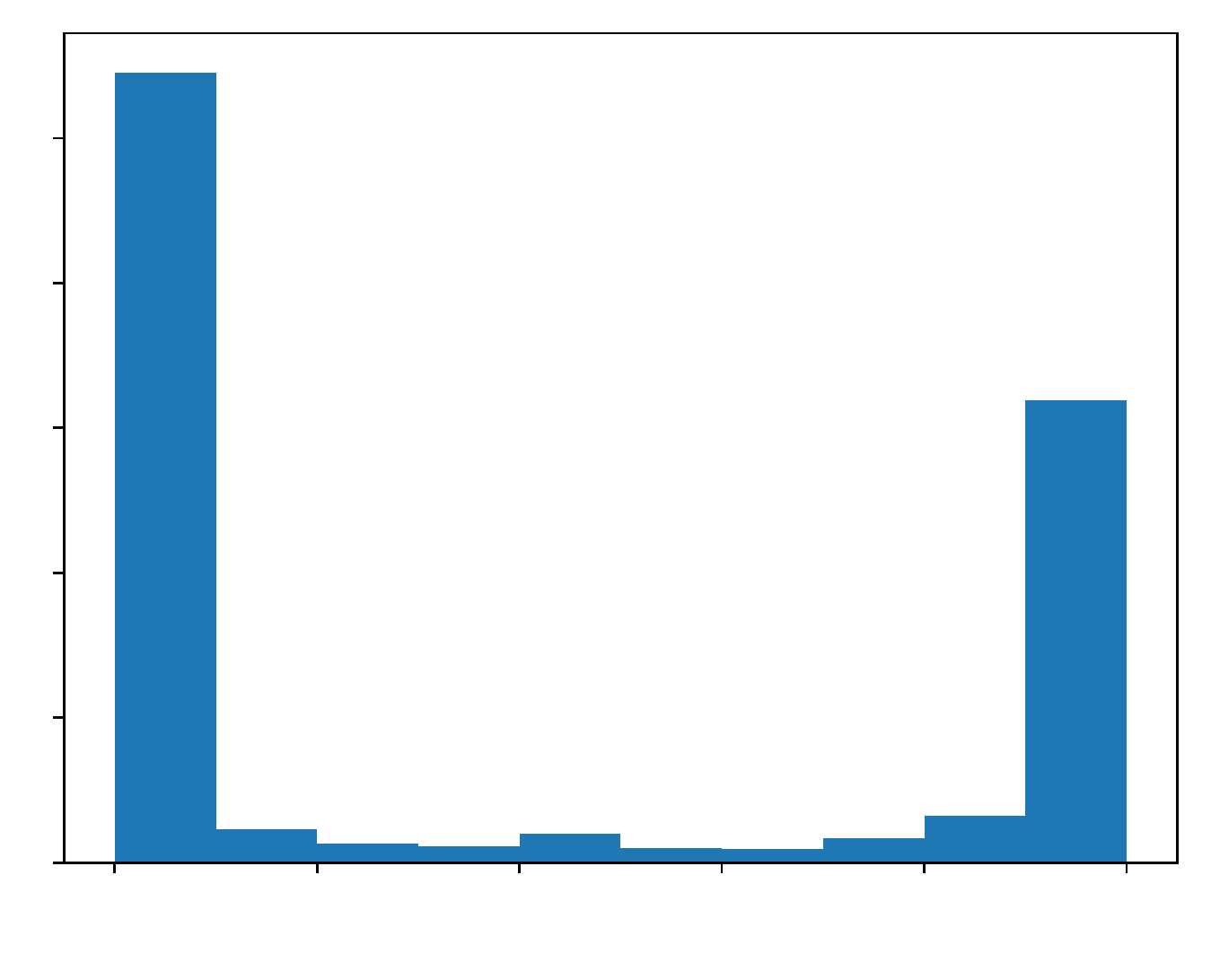}&
		\includegraphics[width=0.13\textwidth]{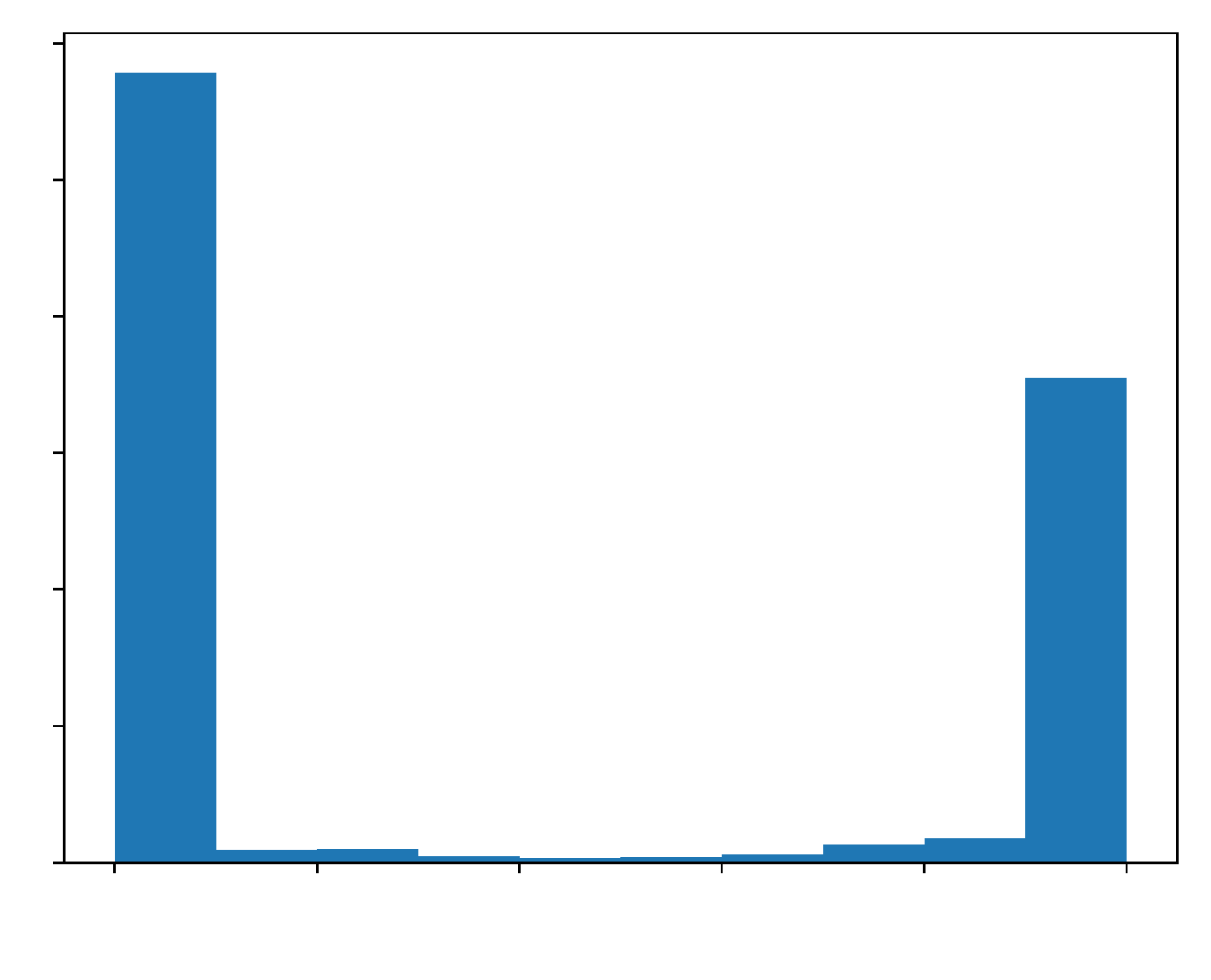} \\
		\multicolumn{6}{c}{Decoder} \\
	\end{tabular}
\caption{Histograms of the domain proportions of each layer in our domain mixing model for English-to-German Task. Within each histogram, $0$ means pure News domain, and $1$ means pure TED domain.} 
\label{fig:domainprop_hist}
\end{figure}

\noindent $\bullet$  The domain proportions of the decoder tend to be more skewed than those of the encoder, which demonstrates little knowledge sharing. Figure~\ref{fig:domainprop_hist} shows the histograms of word-level domain proportions at different layers in both the encoder and decoder. This might explain why the mixing decoder only contributes limited performance gain for the English-to-German task. 


\vspace{-0.05in}
\subsection{Combining Domain Mixing with Domain Aware Embedding}\label{sec:grad_flow}

\begin{figure}
		\centering
	\begin{tabular}{c}
		\includegraphics[width=0.7\textwidth]{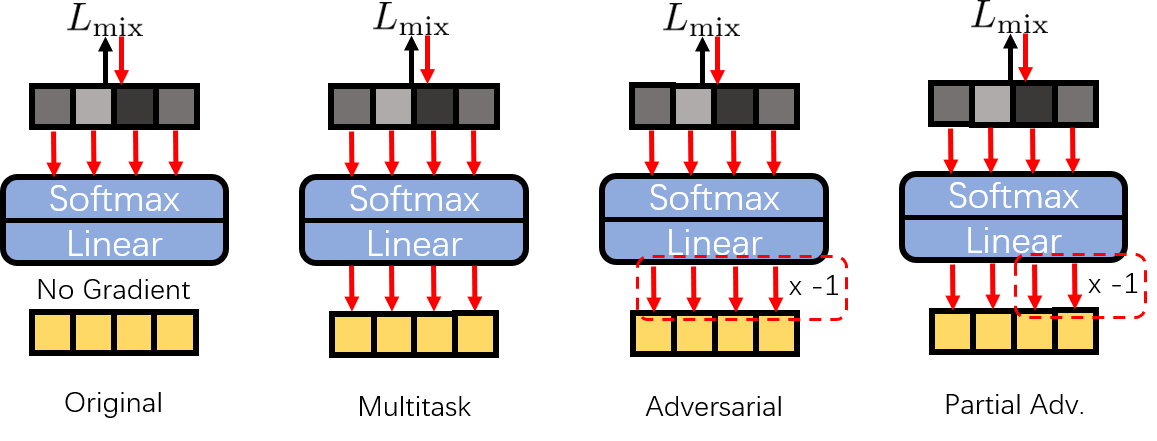}
	\end{tabular}
	\caption{Back-propagation for different embedding based methods.} 
	\label{fig:mix_embed}
\end{figure}

The embedding based methods can be naturally combined with our domain mixing methods. As we mentioned in \ref{sec:implementation_details}, the domain proportion is trained solely, meaning gradient does not propagate between the domain proportion layers $\cD$ and the Transformer. The computation of the gradient, on the other hand, is the key to combining two methods. Specifically, we encourage the embedding to be domain aware via MTL, AdvL and PAdvL, where we use the domain proportion layers to guide the training of the embedding. Figure~\ref{fig:mix_embed} illustrates the back-propagation under different methods. Table~\ref{tab:zh2enExp_mixembed} shows the performance for Chinese-to-English task under this setting. Here we consider applying domain mixing only to the encoder as the baseline. As can be seen, by applying appropriate domain aware embedding, the performance can be further improved.


\begin{table}[!hbt]
	\centering
	\begin{tabular}{c|c|c|c|c}
	    \hline
		Method & Laws & News & Speech & Thesis   \\
		\hline
		\hline
		Encoder & 50.21&	27.94&	16.85&	12.03\\
		+MTL & 49.15 & 26.82 & 15.72 & 11.93 \\
		+Adv &50.18 & 27.72 & \textbf{16.99} & \textbf{12.16}\\
		+PAdvL &49.01	& 26.63	& 16.06	& 12.15 \\
		+Multitask + WL&48.75 & 26.78 & 16.53 & 12.11\\
		+Adv + WL&\textbf{50.24}&	\textbf{28.21}&	16.98&	12.00\\
		+PAdv + WL & 48.87 & 26.86 & 16.14 & 11.89\\
		\hline
	\end{tabular}
	\caption{BLEU Scores of Domain Mixing + Domain Aware Embedding for Chinese-to-English Task} 
	\label{tab:zh2enExp_mixembed}
\end{table}
\vspace{-0.1in}
\section{Discussions} \label{sec:discussion}
\vspace{-0.1in}

One major challenge in multi-domain machine translation is the word ambiguity in different domains. For example, the word ``article'' has different meanings in the domains of laws and media. When translating ``article" into Chinese, the translated words are \begin{CJK*}{UTF8}{gbsn}
	``条款''
\end{CJK*} and \begin{CJK*}{UTF8}{gbsn}
	``文章''
\end{CJK*}, meaning a separate clause of a legal document and a piece of writing. Our proposed word-level layer-wise domain mixing approach tends to reduce the word ambiguity. As mentioned in Section~\ref{sec:layerwise}, our model extracts different representations of each word from contexts at different layers. Accordingly, the domain proportion of each word evolves from bottom to top layers, and can eventually help identify the corresponding domains.

\begin{table}[!htb]
	\centering
	\begin{tabular}{ c|c } 
		\hline
		\multirow{2}{*}{Laws} & ``\textbf{Article} 37 The freedom of marriage ...'' \\
		& \begin{CJK*}{UTF8}{gbsn}
			``第三十七\textbf{条}:婚姻的自由 ...''
		\end{CJK*} \\
		\hline
		\multirow{2}{*}{Media}& ``... working on an \textbf{article} about the poems ...'' \\
		& \begin{CJK*}{UTF8}{gbsn}
			``... 为了一篇诗的\textbf{文章} ...''
		\end{CJK*}\\
		\hline
	\end{tabular}
	\caption{The ambiguity of ``articles''.}
\end{table}

Moreover, as mentioned in Section~\ref{sec:word_level_mixing}, the positional embedding also contributes to the word disambiguation in multi-domain translation. For example, in the law domain, we find that ``article'' often appears at the beginning of a sentence, while in the media domain, the word ``article'' may appear in other positions. Therefore, varying domain proportions for different positions can help with word disambiguation.

We remark that word disambiguation across domains actually requires $\cD(x)$ to be powerful for predicting the domain of the word. However, a powerful $\cD(x)$ tends to yield skewed domain proportions and is not flexible enough for domain knowledge sharing. To trade off between strength and flexibility of $\cD(x)$, the smoothing parameter $\epsilon$ of $\cD(x)$ (see Section~\ref{sec:domainprop}) needs to be properly set.

\vspace{-0.1in}
\section{Conclusions} \label{sec:conclusion}
\vspace{-0.1in}

We present a novel multi-domain NMT with word-level layer-wise domain mixing, which can adaptively exploit the domain knowledge. Unlike the existing work, we construct multi-head dot-product modules for each domain and then combine them by the layer-wise domain proportion of every word. The proposed method outperforms the existing embedding based methods. We also show mixing method can be combined with embedding based methods to make further improvement. 

Moreover, we remark that our approach can be extended to other multi-domain or multi-task NLP problems. We will also investigate how to apply our approach to more complex architectures, e.g., pre-trained BERT encoder \citep{devlin2018bert}. 

\bibliographystyle{ims}
\bibliography{ref}

\clearpage

\appendix

\section{Complementary Experiments} \label{sec:extraexp}

\textbf{Chinese to English}\\
Experiment results of the original Transformer, where layer normalization is at the end each layer.
\begin{table}[!hbt]
    \centering
    \begin{tabular}{c|c|c|c|c}
    Method & Laws & News & Spoken & Thesis   \\
    \hline
    \hline
    Laws &  10.37&  0.45&  0.27&  0.27\\
    News &  0.39&  5.12&  0.91&  0.57\\
    Spoken &  0.70&  1.11&  6.19&  0.83\\
    Thesis &  0.63&  0.25&  0.16&  1.24\\
    Mixed &  5.45&  4.09&  2.67&  1.85\\
    \hline
    Multitask & 6.16& 3.83&  1.91&  1.53\\
    Adversarial  & 5.93& 3.38&  1.85&  1.37\\
    PAdv & 6.58& 3.90&  2.32&  1.80\\
    WDC. w/ WL& 7.13& 3.87& 2.45 & 1.88 \\
    \hline
    \multicolumn{5}{l}{Our proposed Mixing Method}\\
    \hline
     Encoder & 50.16& 27.61&  16.92&  11.85\\
    + Decoder & 50.45& 28.15&  17.45&  11.62  \\
    \end{tabular}
    \caption{Chinese to English}
    \label{tab:zh2enExp}
\end{table}

\begin{figure}[!hbt]
    \centering
    \includegraphics[width=0.7\textwidth]{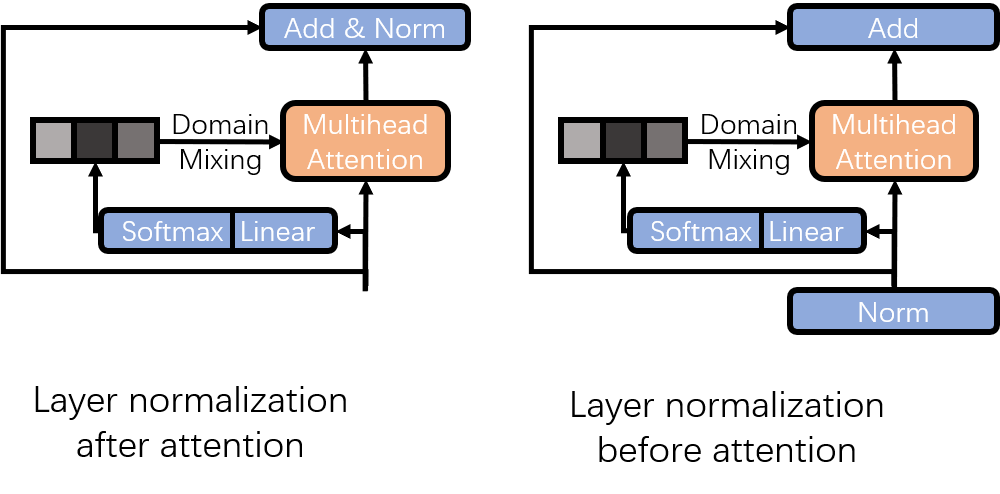}
    \caption{Two variants of layer normalization}
\end{figure}

\end{document}